\documentclass[10pt,twocolumn,letterpaper,table]{article}

\usepackage{cvpr}
\usepackage{times}
\usepackage{epsfig}
\usepackage{graphicx}
\usepackage{amsmath}
\usepackage{amssymb}
\usepackage{url}
\usepackage{bbm}
\usepackage{marvosym}
\usepackage{subfigure}
\usepackage[boxed,linesnumbered]{algorithm2e}
\usepackage{booktabs}
\usepackage{color}
\usepackage{xcolor}
\usepackage{pifont}
\usepackage{flushend}
\usepackage{pbox}

\definecolor{orange}{rgb}{1,0.9,0.65}
\definecolor{gr}{rgb}{0.9,1,0.6}
\definecolor{bl}{rgb}{0.9,0.8,1}
\definecolor{bg}{rgb}{0.8,0.9,1}
\definecolor{orr}{rgb}{1,0.85,0.4}
\definecolor{grr}{rgb}{0.8,1,0.5}
\definecolor{blr}{rgb}{0.85,0.45,1}
\definecolor{bgr}{rgb}{0.5,0.9,1}
\definecolor{gry}{rgb}{0.92,0.92,0.92}
\definecolor{rr}{rgb}{1,0.85,0.9}
\definecolor{blb}{rgb}{0.78,0.84,1}
\definecolor{rdl}{rgb}{0.83,0.39,0.39}
\definecolor{Gray}{gray}{0.75}
\definecolor{Gray2}{gray}{0.55}
\DeclareMathOperator*{\argmax}{\arg\!\max}

\usepackage[pagebackref=true,breaklinks=true,letterpaper=true,colorlinks,bookmarks=false]{hyperref}

 \cvprfinalcopy 

\def\cvprPaperID{611} 

\ifcvprfinal\fi
\begin{document}

\title{A Coarse-to-Fine Model for 3D Pose Estimation and Sub-category Recognition}

\author{Roozbeh Mottaghi$^{1}$\thanks{The work was done while the first author was at Stanford University.}, Yu Xiang$^{2,3}$, and Silvio Savarese$^3$\\
$^1$Allen Institute for AI, $^2$University of Michigan-Ann Arbor, $^3$Stanford University  
}
\maketitle

\begin{abstract}
Despite the fact that object detection, 3D pose estimation, and sub-category recognition are highly correlated tasks, they are usually addressed independently from each other because of the huge space of parameters. To jointly model all of these tasks, we propose a coarse-to-fine hierarchical representation, where each level of the hierarchy represents objects at a different level of granularity. The hierarchical representation prevents performance loss, which is often caused by the increase in the number of parameters (as we consider more tasks to model), and the joint modeling enables resolving ambiguities that exist in independent modeling of these tasks. We augment PASCAL3D+ \cite{xiang_wacv14} dataset with annotations for these tasks and show that our hierarchical model is effective in joint modeling of object detection, 3D pose estimation, and sub-category recognition.
\end{abstract}

\section{Introduction}
Traditional object detectors \cite{viola01, vedaldi09, felzenswalb10} usually estimate a 2D bounding box for the objects of interest. Although the 2D bounding box representation is useful, it is not sufficient. In several applications (e.g., autonomous driving or robotics manipulation), we need to reason about objects' 3D pose or viewpoint in addition to their bounding box location. Therefore, pose estimation methods \cite{thomas06, savarese07, arie09} have been developed to provide a richer description for objects in terms of their viewpoint/pose. Fine-grained recognition methods \cite{farrel11,yao12,berg13} are another class of methods that also aim to provide richer descriptions since they enable more accurate reasoning about the detailed geometry and appearance of objects. Ideally, an object detector should estimate an object's location, its 3D pose and sub-category.

Note that these three tasks, namely object detection, 3D pose estimation, and sub-category recognition, are correlated tasks. For instance, learning an object model for \textit{sedans} seen from a particular viewpoint is `easier' than learning a model for general \textit{cars} as the former forms a tighter cluster in the appearance space. On the other hand, more accurate localization of the object helps to better estimate its sub-category and viewpoint. Although these tasks are highly correlated, they are usually solved independently. One of the main issues in joint modeling of these tasks is that the number of parameters increases as we consider more tasks to model. This typically leads to requiring a larger number of images for training in order to avoid overfitting and performance loss compared to independent modeling. For instance, images of a particular type of \textit{truck} taken from a certain viewpoint might be rare in the training set, hence learning a robust model for that might be difficult. This issue has been addressed in the literature by different techniques (for example, part sharing between 
different viewpoints \cite{hejrati12, xiang12}). In this work, we take an alternative approach and leverage coarse-to-fine modeling.

\begin{figure}
\centering
  \includegraphics[width=18pc]{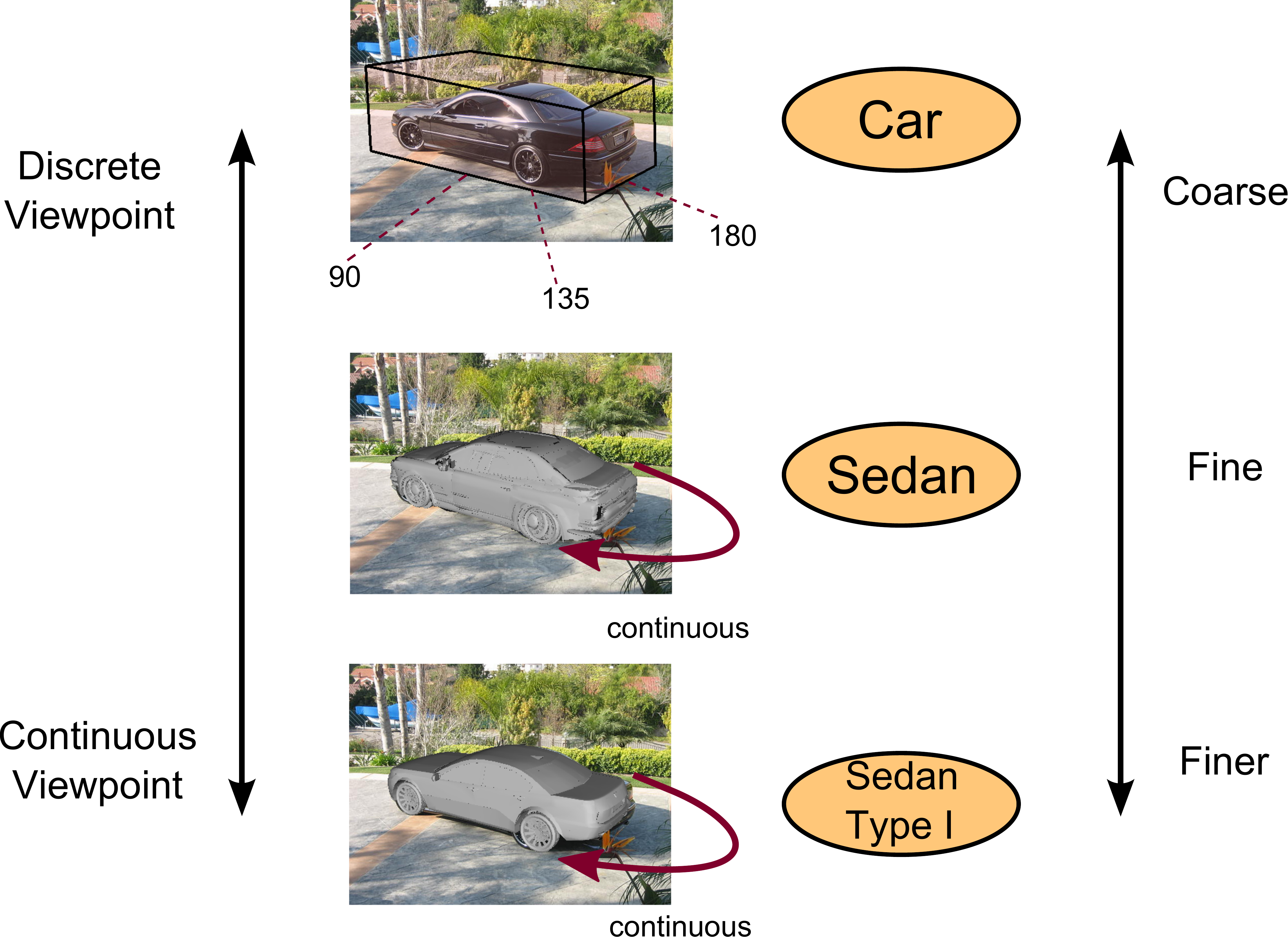}
  \caption{A coarse-to-fine hierarchical representation of an object. The top-layer captures high-level information such as a discrete viewpoint and a rough object location, while the layers below represent the object more accurately using continuous viewpoint, sub-category, and finer-sub-category information.\vspace{-0.5cm}}
  \label{fig:intro}
\end{figure}

We propose a novel coarse-to-fine hierarchical model to represent objects, where each layer of the hierarchy represents objects at a different level of granularity. As shown in Figure~\ref{fig:intro}, the coarsest level of the hierarchy reasons about the basic-level categories (e.g., \textit{cars} vs. other categories) and provides a rough discrete estimate for the viewpoint. As we go down the hierarchy, the level of granularity changes, and more details are added to the model. For instance, for \textit{car} recognition, at one level we reason about sub-categories such as \textit{SUV}, \textit{sedan}, \textit{truck}, etc., while at a finer level we distinguish different types of \textit{SUVs} from each other. Also, we have a more detailed viewpoint representation (continuous viewpoint) in the layers below. 

There are advantages of this coarse-to-fine hierarchical representation. First, tasks at different levels of granularity can benefit from each other. For instance, if there is ambiguity about the viewpoint of the object, knowing the sub-category might help resolving the ambiguity or reduce the uncertainty in viewpoint estimation. Second, different types of features are required for these three tasks. For instance, a feature that is most discriminative for distinguishing \textit{cars} from other categories is not necessarily useful for distinguishing different types of \textit{SUVs}. The hierarchical representation provides a principled framework to learn feature weights for different tasks jointly. Finally, we can better leverage the structure of the parameters so the performance does not drop as we increase the complexity of the model (or equivalently, the layers of the hierarchy).

Our hierarchical model is a hybrid random field as it contains discrete (e.g., sub-category) and continuous (e.g., continuous viewpoint) random variables. We employ a particle-based method to handle the mixture of continuous and discrete variables in the model. During learning, the parameters of the model in all layers of the hierarchy are estimated jointly. Inference is also a joint estimation of the object location, and its continuous viewpoint, sub-category and finer-sub-category. 

For our experiments, we use PASCAL3D+ \cite{xiang_wacv14} dataset, which provides viewpoint annotations for rigid categories of PASCAL VOC 2012 dataset. To evaluate and train our model, for a subset of categories, we augment PASCAL3D+ with sub-category and finer-sub-category annotations. Our results show that our hierarchical model is effective in joint estimation of object location, 3D pose and (finer-)sub-category information. Also, the performance typically does not drop significantly or even improves as we increase the complexity of the model. Moreover, the hierarchical model provides significant improvement over a flat model that uses the same set of features.

\section{Related Work}
\textbf{Hierarchical Models.} Hierarchical models have been used extensively for object detection and recognition. \cite{sanja07} and \cite{zhu08} use hierarchies of object parts for object detection, where the parts in each layer are a composition of the parts in layers below. \cite{sivic08} discover a hierarchical structure to group objects based on common visual elements. \cite{salakhutdinov11} uses a hierarchy to share features between categories so they boost the recognition performance for categories with few training examples. We use a hierarchy as a unified model for 3D pose estimation, sub-category recognition, and object detection. The motivation, representation and the details of our model are different from the mentioned methods.


\textbf{3D Pose Estimation.} Several methods address the problem of object detection and pose estimation by incorporating 3D cues. Here we mention a few examples. Some of these methods, such as \cite{su09,payet11}, link parts across views, which allows a continuous viewpoint representation. \cite{liebelt10,hejrati12} treat 2D appearance and 3D geometry separately and combine them in a later stage. Hedau et al. \cite{hedau10} represent object appearance by a rigid template in 3D. Fidler et al. \cite{fidler12} extend that work by considering deformable faces. The methods mentioned above are limited to basic-level categorization, while we reason about sub-category information as well. 

\textbf{Sub-category Recognition.} There is a considerable body of work on fine-grained categorization in the 2D recognition literature \cite{farrel11,yao12,berg13,duan12,parkhi12}, which typically ignore reasoning about the 3D information. Recently, the 3D recognition community has shown that 3D object representation is beneficial for fine-grained categorization and vice versa. The work by \cite{zia13} infers sub-categories in addition to the 3D pose. However, their sub-category recognition is performed as a post-processing step, while we perform that in a joint fashion. \cite{lim13} also address the problem of viewpoint and sub-category estimation. However, they solve a binary classification problem (a particular sub-category vs. background), while we solve a multi-class problem, which is more challenging. \cite{stark12} uses fine-grained category information to better understand a scene in 3D. \cite{krause13} extends Spatial Pyramid Matching and Bubble Bank to 3D to perform fine-grained categorization and viewpoint estimation. \cite{lin14} optimize fine-grained recognition and 3D model fitting jointly. \cite{pepik14} propose a transfer learning method for simultaneous object localization and viewpoint estimation and show that this transfer is beneficial for sub-category estimation. These methods suffer from one or more of the following issues. They assume the object bounding box is given, work only on clean images that do not contain any occlusion, cannot estimate continuous viewpoint or cannot estimate elevation of the camera or its distance from the object.

\begin{figure*}[ht]
\begin{minipage}[b]{0.65\linewidth}
\centering
\includegraphics[width=\textwidth]{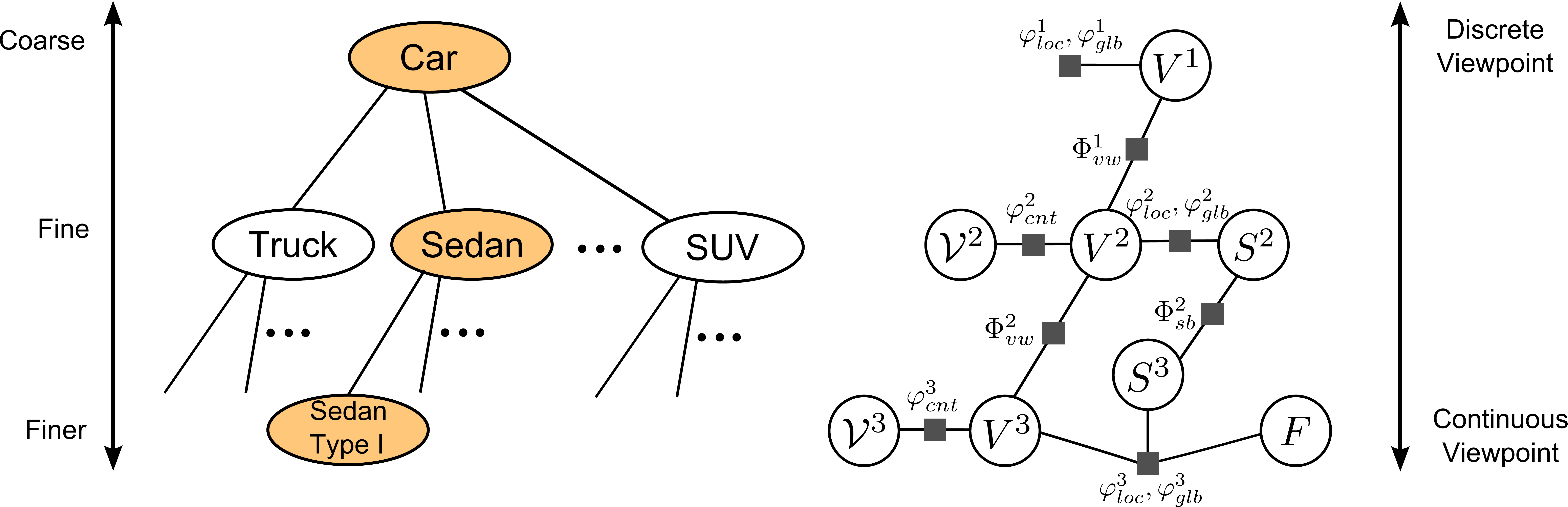}
\caption{The graphical model of the hierarchy. For clarity, we have removed object node $O$. On the squares we have shown the potential functions defined on the nodes connecting to them. See text for the details.}
\label{fig:model}
\end{minipage}
\hspace{0.5cm}
\begin{minipage}[b]{0.25\linewidth}
\centering
\includegraphics[width=\textwidth]{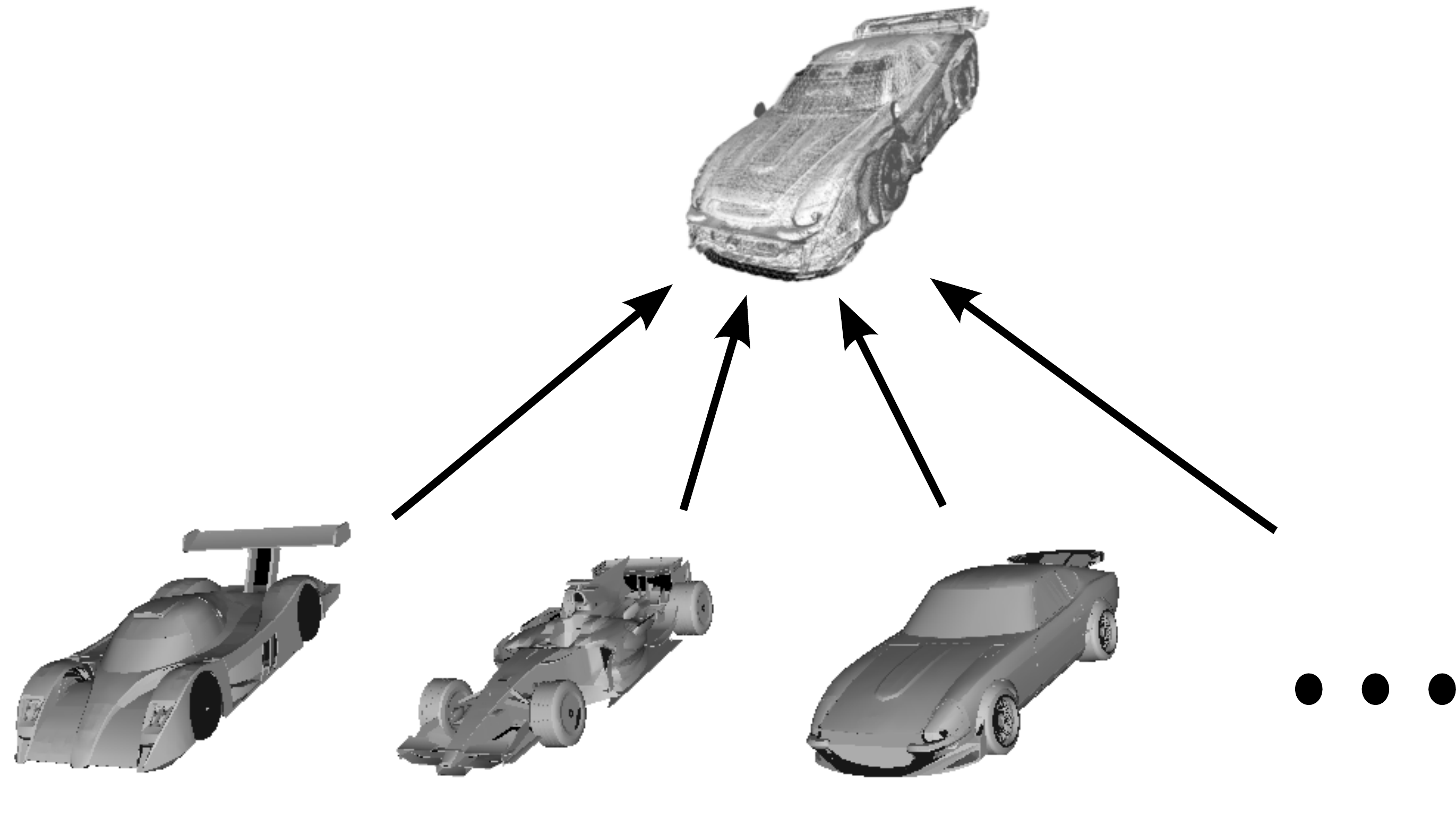}
\caption{A coarse CAD model is made from the more detailed CAD models in the layers below. See text for more details.}
\label{fig:cads}
\end{minipage}
\end{figure*}

\section{Coarse-to-fine Hierarchical Object Model}
In this section, we describe our hierarchical model, which jointly performs object detection, 3D pose estimation, and sub-category recognition. The key intuition is that an object can be represented at different levels of granularity in different layers, where some constraints impose consistency across layers. We formulate the problem as learning and inference in a hybrid random field, which contains a mixture of discrete and continuous random variables. The hierarchy that we consider has three layers. The top layer (coarsest layer) captures coarse information, i.e., the object label (e.g., \textit{aeroplane} or not) and also a coarse (discretized) viewpoint. This information is represented by a set of discrete random variables. The layer below in the hierarchy adds information about sub-category (e.g., \textit{airline aeroplane}, \textit{fighter aeroplane}, \textit{etc.}) and also continuous viewpoint. Sub-category is represented by a discrete variable, while a continuous random variable 
represents the continuous viewpoint information. The bottom layer (or the finest layer) adds detailed information about the sub-categories that we refer to as finer-sub-category (e.g., a certain type of \textit{airline aeroplane}). Viewpoint information is represented using a continuous random variable at this layer as well.

More formally, the binary random variable $O$ represents the object label, where it will be equal to 1 if it is the object of interest and 0 otherwise. The coarse viewpoint is denoted by $V^l$, which takes values in the following discrete set of coarse viewpoints $\mathcal{A} = \{a_1,a_2,\ldots,a_m, b\}$, where $m$ specifies the number of azimuth sections, and $b$ represents background (no viewpoint should be associated to a background region). Therefore, each section covers $360 / m$ degrees. The superscript $l$ indexes the level in the hierarchy. The continuous viewpoint is denoted by $\mathcal{V}^l = (a,e,d,occ)$, which is decomposed into azimuth $a$, elevation $e$, distance (depth) $d$, and occlusion $occ$. We will describe these variables in more detail when we describe the potential functions defined on them. Another variable in the model is the sub-category variable $S^l$, which chooses a value from the set $\mathcal{S} = \{s_1,s_2,\ldots,s_n, b\}$, where $n$ is determined according to the number of sub-categories we consider for an object category. Similarly, the random variable $F$ represents the finer-sub-category in the model and selects a label in the set $\mathcal{F}_s=\{f_{s1}, f_{s2}, \ldots, f_{sp}, b\}$, where $s$ indexes the subcategories and $p$ indexes the finer-sub-categories of sub-category $s$. 

\subsection{Potential functions}
\label{sec:pot}
We now describe the potential functions defined for our three layer hierarchy. The level of the potential function is specified by the superscript $l$, e.g.,  $\varphi^l_.$. We have illustrated the graphical model for object $O$ in Figure~\ref{fig:model}.

\noindent\textbf{Global shape.} We capture the global shape of the objects with HOG templates. We denote these potential functions as $\varphi^1_{glb}(V^1;\mathcal{R})$, $\varphi^2_{glb}(V^2,S^2;\mathcal{R})$, and $\varphi^3_{glb}(V^3,S^3,F;\mathcal{R})$. As mentioned above, $V^l$ corresponds to the viewpoint and $S^l$ and $F$ denote the (finer-)sub-category information. Note that the term in the first layer of the hierarchy is a function of the viewpoint only, while in the layers below, it becomes a function of viewpoint and sub-category. These terms basically represent the HOG feature that we compute for region $\mathcal{R}$. Region $\mathcal{R}$ is a proposal bounding box in the image, which can be generated by methods such as \cite{jasper13}.

\noindent\textbf{Local appearance.} We introduce these terms to capture local appearance information. For this purpose, we train a convolutional neural network (CNN) to compute the features used in the potential functions. We refer to them as `local', because typically CNN units respond on portions of the objects and implicitly act as a `part detector'. We use the CNN implementation of \cite{girshick14}, but use only five convolutional layers to compute the features. We denote these terms by $\varphi^1_{loc}(V^1;\mathcal{R})$, $\varphi^2_{loc}(V^2,S^2;\mathcal{R})$, and $\varphi^3_{loc}(V^3,S^3,F;\mathcal{R})$ for the three layers of the hierarchy. Similar to above, the CNN features are computed on region $\mathcal{R}$.

\noindent\textbf{Continuous viewpoint.} The terms defined so far are based on a discretized viewpoint (discrete azimuth angle only). The azimuth angle alone is not sufficient to accurately represent the 3D pose of an object. This term in the energy function is computed based on the alignment of image data with the projection of a 3D CAD model. An advantage of using the 3D CAD models is that we can search for viewpoints not observed during training since the CAD models can be rendered from any viewpoint and also we can better reason about occlusions with 3D CAD models.

\begin{figure}[tp]
\centering
\subfigure[]{
   \includegraphics[width=12pc] {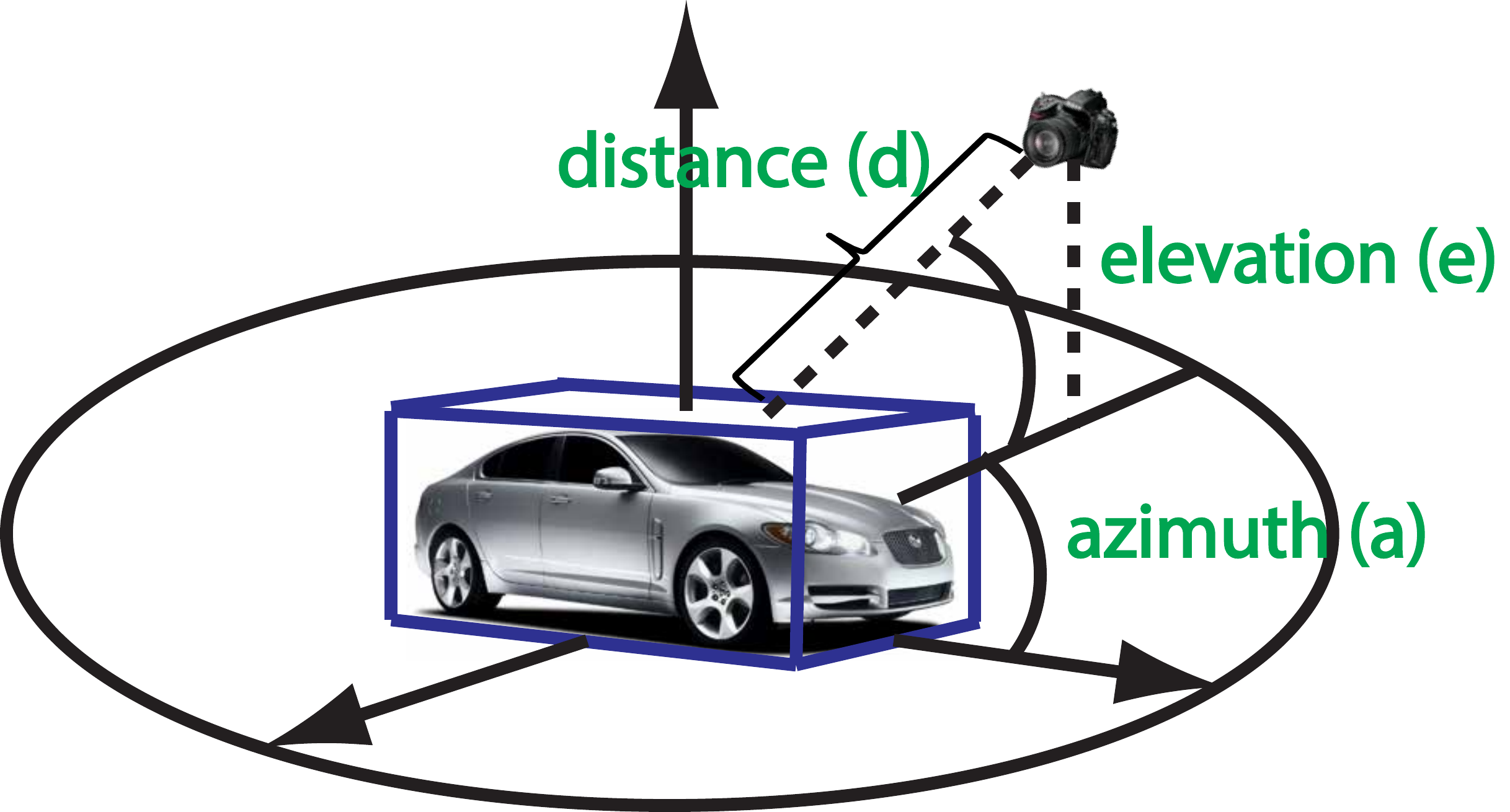}
   \label{fig:aed}
 }
 \subfigure[]{
   \includegraphics[width=14pc] {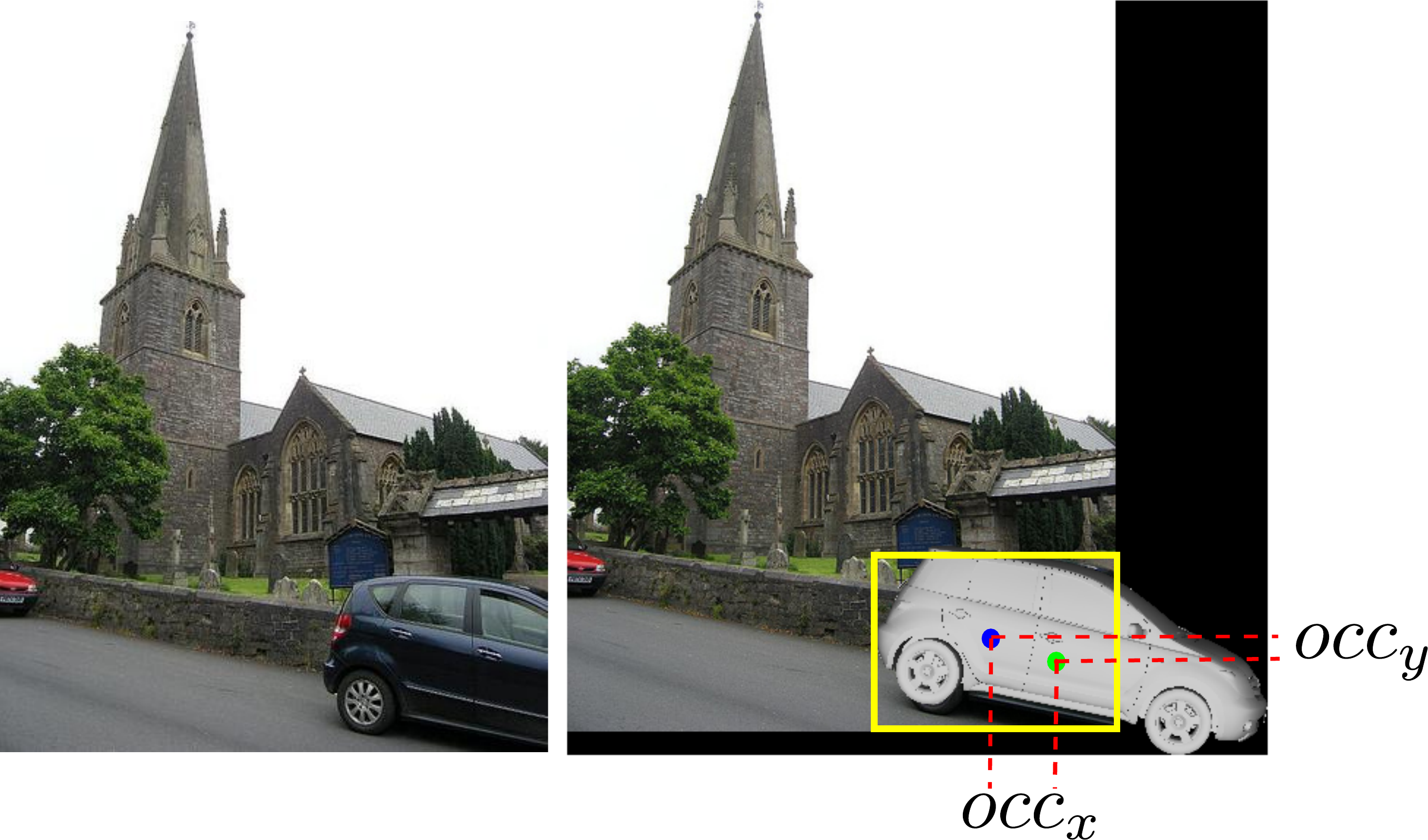}
   \label{fig:occ}
 }
 \caption{Parameters of the continuous viewpoint.}
\label{fig:contview}
\end{figure}

The potential function that we now define makes the connection between the continuous variable $\mathcal{V}^l$, which denotes the continuous viewpoint, and the discretized viewpoint $V^l$. The continuous viewpoint is a 4-tuple $\mathcal{V}^l = (a,e,d,occ)$. The range of azimuth angle $a$ is $[0,2\pi)$, while the elevation angle $e$ is in the range $[0,\pi/2]$. The distance (depth) $d$ corresponds to the distance of the camera from the object. The 3D pose of an object can be determined by these three parameters. For clarification, we show these parameters in Figure~\ref{fig:aed}. The last variable $occ$ is for better handling of  truncation and occlusion and it is described below. 

The idea for using the occlusion variable $occ$ is that we translate the projected CAD model in a neighborhood around the original point of projection (center of the bounding box), so it better fits the observation in the image. For instance, in Figure~\ref{fig:occ}, if we translate the projection of the CAD model to the right, it will be better aligned with the truncated car. Basically, $occ$ is a translation vector that moves the projection from the center of the bounding box (blue point) to a new location (green point).

The alignment between the projection of the CAD model and the observation in the image is computed as follows. We render the 3D CAD model onto the image according to $\mathcal{V}^l$. Then we compute HOG features on the contour (outline) of the projection and compare it with the HOG feature computed on region $\mathcal{R}$. We consider only the portion of projection that falls into $\mathcal{R}$. 

The potential function is defined as:
\begin{equation}
\varphi^l_{cnt}(V^l, \mathcal{V}^l, C^l; \mathcal{R}) = \frac{1}{|\mathcal{R}|} \max_{\nu^l} \phi(P_{\nu^l, C^l})^T\phi(\mathcal{R}),
\label{eq:cont}
\end{equation}
where $\phi(.)$ denotes the HOG feature and $P_{\nu^l, C^l}$ is the projection of the CAD model, $C^l$, according to $\nu^l$. We perform normalization so this term does not depend on the scale of $\mathcal{R}$. $\nu^l$ is a set of samples that are generated according to the discrete viewpoint, and the one that maximizes the alignment between $\phi(P_{\nu^l, C^l})$ and $\phi(\mathcal{R})$ (described above) is chosen to compute the potential function. The samples of the continuous viewpoint variable are generated as follows: $\nu^l_a \sim \mathcal{N} (v^l; \sigma_a)$, $\nu^l_e \sim \mathcal{N} (\mu_e; \sigma_e)$,  $\nu^l_{occ} \sim \mathcal{N} (\mathcal{R}_c; \sigma_{rx}, \sigma_{ry})$, where $\nu ^l_a$, $\nu^l_e$, and $\nu^l_{occ}$ represent azimuth, elevation and the occlusion variable in the continuous viewpoint, respectively. $v^l$ is one of the $m$ discrete values in $\mathcal{A}$ (recall that the discrete viewpoint is only defined on the azimuth angle), $\mu_e$ is the average of elevations in training data, and $\mathcal{R}_c$ is the center of the proposal bounding box. We empirically set $\sigma_a$ and $\sigma_{r.}$, and $\sigma_e$ is computed from training data. 

This sampling strategy allows us to make a connection between the continuous and discrete viewpoints. Note that solving for unconstrained continuous variables directly is difficult. The discrete variables somewhat constrain the values that the continuous variables can take. Furthermore, computing the right hand side of Equation~\ref{eq:cont} requires maximization over a continuous domain, which is not practical. Sampling makes this problem tractable as well.

\begin{figure}[h]
\centering
  \includegraphics[width=12pc]{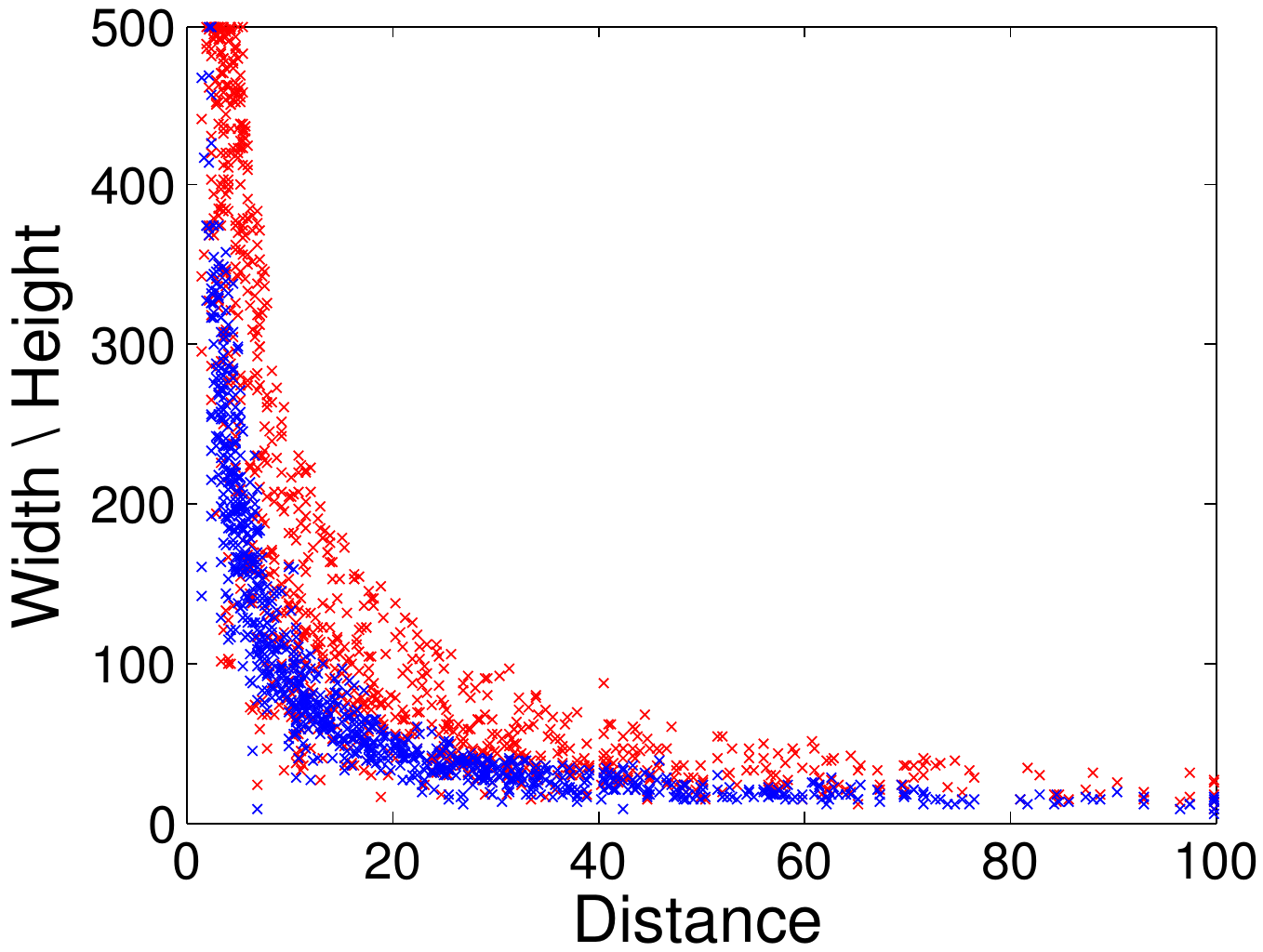}
  \caption{Correlation of object distance with the height and width (in pixels) of its 2D bounding box for \textit{car} training instances. Width is shown in red and height in blue.}  
  \label{fig:hw}
\end{figure}

The distance $d$ is sampled differently from the other parameters. We use the following simple procedure for sampling the distance, but more sophisticated methods can be adopted instead. As shown in Figure~\ref{fig:hw}, there is a correlation between distance $d$ and size of the proposal box $\mathcal{R}$. During training, we know both distance and box size. During test, we have to estimate the distances given the proposal box size. We assign a weight to each training instance based on the difference in width and height of the training instances and the test instance (higher weight to smaller differences). We sample training instances according to these weights and use their distance $d$ to form the set of distance samples. 

A small proposal bounding box can correspond to a far away object or it can correspond to a nearby but truncated/occluded object. The distance sampling enables us to explore both of these possibilities.


Now, the question is which 3D CAD model, $C^l$, should be selected for computing this term. For the bottommost layer of the hierarchy, we collect different CAD models to represent intra-class variation in a sub-category. For the mid layer, we combine the fine-grained CAD models in the lower layer to make a new CAD model, which captures generic shape properties of the object sub-category. For instance, we combine all different types of race cars to make a coarse race car model (Figure~\ref{fig:cads}). To combine the CAD models we scale them to the same size and orient them to a common direction. Then, we superimpose the CAD models and voxelize them. We keep only the voxels that vertices from a certain fraction of the CAD models fall into them. 

\noindent \textbf{Across layer consistency.} To impose consistency between different layers we define a set of pairwise potentials. The discrete viewpoint should be the same across all layers. Also, the sub-category should be consistent across layers. So,
\begin{align}
&\Phi^l_{vw}(V^l, V^{l+1}) = \begin{cases} 1 & v^l=v^{l+1} \\ -\infty & \mbox{otherwise}\end{cases}\,\,\,\, l = 1,2  \\ &\Phi^l_{sb}(S^l, S^{l+1}) = \begin{cases} 1 & s^l=s^{l+1} \\ -\infty & \mbox{otherwise}.\end{cases}\,\,\,\, l = 2 
\end{align}

Note that we do not enforce direct consistency between continuous viewpoints, as they might be different depending on the level of granularity of the CAD model. 

\noindent \textbf{Top-level Detector.} We use a pre-trained binary classifier that is applied to the proposal boxes and determines the confidence of a box belonging to the basic-level category of interest. In particular, we use the classifier of \cite{girshick14}. We denote this potential function by $\varphi_{det}(O;\mathcal{R})$. 

\subsection{Full energy function}
The energy function is written as the sum of the energy functions in the three layers of the hierarchy:
\begin{multline}
E = \sum_{l=1}^3 E^l = w_1 \varphi_{det} + \sum_{l=1}^3 \big( {\mathbf{w}_2^l}^T \varphi^l_{glb} + {\mathbf{w}_3^l}^T \varphi^l_{loc} \big) + \\
\sum_{l=2}^3 {\mathbf{w}_4^l}^T \varphi^l_{cnt} + \sum_{l=1}^2 {\mathbf{w}_5^l}^T \Phi^l_{vw} +{\mathbf{w}_6^l}^T \Phi^2_{sb}, 
\end{multline}
where $\mathbf{w}$'s are the parameters of the model that are estimated by the learning method described below.

\section{Learning \& Inference}
As the result of inference on our model we can determine if a proposal box belongs to the category of interest and we also estimate its 3D viewpoint, sub-category, and finer-sub-category. Therefore, we find the configuration that maximizes $E(O, \{V^l\}, \{\mathcal{V}^l\}, \{S^l\}, F; \mathcal{R})$ given the weights $\mathbf{w}$ that are estimated during learning:
\begin{multline}
\big(O^*, \{{V^*}^l\}, \{\mathcal{V^*}^l\}, \{{S^*}^l\}, F^*\big) = \\
\argmax_{O, \{{V}^l\}, \{\mathcal{V}^l\}, \{{S}^l\}, F} E(O, \{{V}^l\}, \{\mathcal{V}^l\}, \{S^l\}, F; \mathcal{R}),
\end{multline}
where $l=1,2,3$ for $V^l$, and $l=2,3$ for $\mathcal{V}^l$ and $S^l$.

Our inference method should estimate continuous and discrete variables in the model so we adopt an inference procedure that shares similarities with particle convex belief propagation (PCBP) \cite{peng11}. The continuous variable in the model corresponds to the continuous viewpoint. First, we draw multiple samples around each discrete viewpoint. Basically, these samples can be considered as labels in a discretized MRF and allow us to compute the potential function defined in Eq.~\ref{eq:cont}. After this step, the model can be considered as a fully discrete MRF and we can apply inference techniques for discrete MRFs. The advantage of particle methods is that they prevent committing to a fixed quantization of the state space. We can perform exact inference using exhaustive search since the number of possibilities is not too huge. 

We use a structured SVM framework \cite{ssvm04} to learn the weights in the model. Our positive training examples are a set of bounding boxes for the category of interest. In addition, we provide viewpoint as well as sub-category and finer-sub-category annotations for each example. The loss function $\Delta^l$ depends on the level of the hierarchy as well. We use $\Delta^1$ to penalize mis-prediction of the viewpoints. $\Delta^2$ penalizes sub-category mis-predictions and $\Delta^3$ assigns a penalty to the incorrect predictions of the finer-sub-category. We perform loss augmented inference to find the most violating constraint. Note that each layer contributes its corresponding loss to the total loss. We use the 1-slack cutting plane implementation of \cite{desai11} for the optimization. The details of the learning procedures are summarized in Algorithm~\ref{alg:ssvm}.

 \begin{algorithm}\label{alg:ssvm}\scriptsize
 \SetKwInOut{Input}{input}
 \SetKwInOut{Output}{output}
 \caption{SSVM for our MRF, which is a mixture of continuous and discrete random variables.}
 \Input{Training examples: $\mathbf{x}_i=(o, v, \nu, s, f;\mathcal{R})\,\,\,\,i=1,\ldots,N$}
 \Output{Estimated weights $\mathbf{w}_j$}
 \BlankLine
 Initialize weights $\mathbf{w}_j$ randomly\;
 \For{$t\leftarrow 1$ \KwTo \# of iterations}
 {
   \ForEach{training sample $\mathbf{x}_i$}
   {
     \ForEach{layer $l$}
     {
       Compute the potentials defined based on the discrete variables: $\varphi_{det}, \varphi_{glb}^l, \varphi_{loc}^l, \Phi^l_{vw}, \Phi^l_{sb}$ \;
       \ForEach{possible discrete viewpoint $v\in\mathcal{A}$}
       {
         Sample $K$ continuous viewpoints $\nu$ (according to the sampling strategy in Section 3.1)\;
         \ForEach{sub-category or finer-sub-category (depending on the layer)}
          {
             Project the corresponding CAD model according to the sampled viewpoints\;
					   Compute the corresponding entry in $\varphi_{cnt}^l$\;
          }
       }
     
       Compute the loss function $\Delta^l$ (defined in Section 4)\;     
     }
     Perform loss augmented inference to find the most violating constraint\;
     Solve for $\mathbf{w}_j$ similar to the discrete SSVM\;  
   }   

 }
 \end{algorithm}

\begin{table*}[t]
{\scriptsize
\begin{center}
\addtolength{\tabcolsep}{0.4pt}
\begin{tabular}{|l|c|c|c|c|c|}
\hline
& \textbf{Bounding Box} & \textbf{All} & \textbf{Sub-category \& Viewpoint} & \textbf{Sub-category} & \textbf{Viewpoint} (8 views)\\
\hline
\textbf{RCNN} \cite{girshick14} & 51.4 & \ding{56} & \ding{56} & \ding{56} & \ding{56} \\
\hline
\hline
\textbf{DPM-VOC+VP} \cite{pepik12a} & 29.5 & \ding{56} & \ding{56} & \ding{56} & 21.8 \\
\hline
\hline
\textbf{V-DPM} \cite{felzenswalb10} & 27.6 & \ding{56} & \ding{56} & \ding{56} & 16.2 \\
\hline
\textbf{SV-DPM} \cite{felzenswalb10} & \textbf{27.8} & \ding{56} & \textbf{8.4} & \textbf{13.8} & \textbf{18.2} \\
\hline
\textbf{FSV-DPM} \cite{felzenswalb10} & 25.8 & 0.35 & 7.9 & 12.7 & 16.1 \\
\hline
\end{tabular}
\end{center}
}
\caption{Results of variation of DPM \cite{felzenswalb10}, DPM-VOC+VP \cite{pepik12a} and RCNN \cite{girshick14} on PASCAL3D+ \cite{xiang_wacv14} for all three or a subset of tasks.The result of DPM-VOC+VP \cite{pepik12a} is adopted from \cite{xiang_wacv14}. The first column (`Bounding Box') is equivalent to the standard detection AP of PASCAL VOC. The meaning of \ding{56} is that the method is not capable of doing that task. We have shown the results averaged over classes.}
\label{tab:DPM}
\end{table*}
\begin{table*}[tp]
{\scriptsize
\begin{center}
\addtolength{\tabcolsep}{0.4pt}
\begin{tabular}{|l|c|c|c|c|c|}
\hline
& \textbf{Bounding Box} & \textbf{All} & \textbf{Sub-category \& Viewpoint} & \textbf{Sub-category} & \textbf{Viewpoint} (8 views)\\
\hline
\textbf{1-layer hierarchy (ours)} & 49.5 & \ding{56} & \ding{56} & \ding{56} & 28.9 \\
\hline
\textbf{2-layer hierarchy (ours)} & 51.0 & \ding{56} & 16.0 & 27.5 & \textbf{29.5} \\
\hline
\textbf{3-layer hierarchy (ours)} & \textbf{51.6} & \textbf{3.2} & \textbf{17.6} & 30.6 & \textbf{29.5} \\
\hline
\textbf{Flat model (ours)} & $51.6^\dagger$ & 2.6 & 14.8 & 27.8 & 26.3 \\
\hline
\textbf{Separate (ours)} & $51.6^\dagger$ & 1.9 & 16.1 & \textbf{31.0} & 28.7 \\
\hline
\end{tabular}
\end{center}
}
\caption{Results of variations our hierarchical model, a flat model that uses the same set of features as those of the 3-layer hierarchy, and also separate classifiers on PASCAL3D+ \cite{xiang_wacv14}. $^\dagger$ We consider the same confidence values as those of the 3-layer model. So the bounding box detection results are identical.}
\label{tab:ours}
\end{table*}
\section{Experiments}

In this section, we demonstrate the result of our method for object detection, 3D pose estimation, and (finer-)sub-category recognition. 

\noindent \textbf{Dataset.} For our experiments, we use PASCAL3D+ \cite{xiang_wacv14} dataset, which provides continuous viewpoint annotations for 12 rigid categories in PASCAL VOC 2012. We augment three categories (\textit{aeroplane}, \textit{boat}, \textit{car}) of PASCAL3D+ with sub-category and finer-sub-category annotations. We consider 12, 12, and 60 finer-sub-categories for \textit{aeroplane}, \textit{boat}, and \textit{car} categories, respectively. We group finer-sub-categories into 4, 4, and 8 sub-categories, respectively. For instance, the sub-categories we consider for \textit{cars} are \textit{sedan, SUV, truck, race}, etc., and the finer-sub-categories represent different types of \textit{sedans} or \textit{SUVs}. For the full list, refer to the supplementary material. For each finer-sub-category, we have a corresponding 3D CAD model, and for annotation we assign the instance in the image to the most similar CAD model. We use the \texttt{train} subset of PASCAL VOC 2012 for training, and the \texttt{val} subset for evaluation.  

\noindent \textbf{Implementation details.} For generating proposal bounding boxes ($\mathcal{R}$) we use the method of \cite{jasper13}, but any other method that produces object hypotheses can be used. The losses for the top layer ($\Delta^1$) and the finest layer ($\Delta^3)$ are set to 0.1, and the mid-layer loss ($\Delta^2$) is set to $0.3 / K$, where $K$ is the frequency of the sub-category in training data. The standard deviations used for sampling in Eq.~\ref{eq:cont} is computed as follows. $\sigma_a$ is 1/3 of each azimuth section, $\sigma_e$ is computed from training data, and $\sigma_{r.}$ is set to $0.15\times L$, where $L$ is the maximum of height and width of the proposal bounding box. We compute 5, 3, 2, 2 samples for azimuth, elevation, distance, and $occ$, respectively so we have 60 viewpoint samples in total. We set the $C$ parameter of the structured SVM to 1. The inference takes about a minute per image on a single 3.0 GHz CPU. Most time is used to compute $\varphi_{cnt}^l$ that requires rendering CAD models.

\noindent \textbf{Results.} We evaluate the three tasks using an evaluation method similar to average viewpoint precision (AVP) of \cite{xiang_wacv14}: we consider a box to be correct if the bounding box has more than $50\%$ overlap with ground truth (the standard PASCAL detection criteria), and its viewpoint, sub-category, and finer-sub-category are estimated correctly as well. Therefore, the tasks are much more difficult than the standard bounding box localization. In the tables we show results for all tasks (referred to as `All') as well as a subset of tasks. For example, for evaluating `Sub-category \& Viewpoint', we ignore if the finer-sub-category has been estimated correctly or not. 

We report results for the tasks using various baseline methods. The first is RCNN \cite{girshick14} (refer to Table~\ref{tab:DPM}). For per-class results, refer to the supplementary material. Next we show the results of variations of DPM \cite{felzenswalb10} in Table~\ref{tab:DPM}. V-DPM refers to the case that DPM mixture components correspond to different viewpoints (8 azimuth angles in this case). SV-DPM is the scenario that the mixture components represent both viewpoint and sub-categories (e.g., for \textit{cars}, we consider $8\, \mbox{(viewpoints)} \times 8\, \mbox{(sub-categories)} = 64$ components). Similarly, FSV-DPM considers finer-sub-categories as well (e.g., 60 finer-sub-categories for cars). Our purpose for providing these results is to illustrate the performance drop in all tasks when we compare the results of SV-DPM and FSV-DPM, which is due to the increase in the number of parameters or lack of training instances per component. 

\begin{figure*}[tp]
\vspace{-0.8cm}
\centering
  \includegraphics[width=37pc]{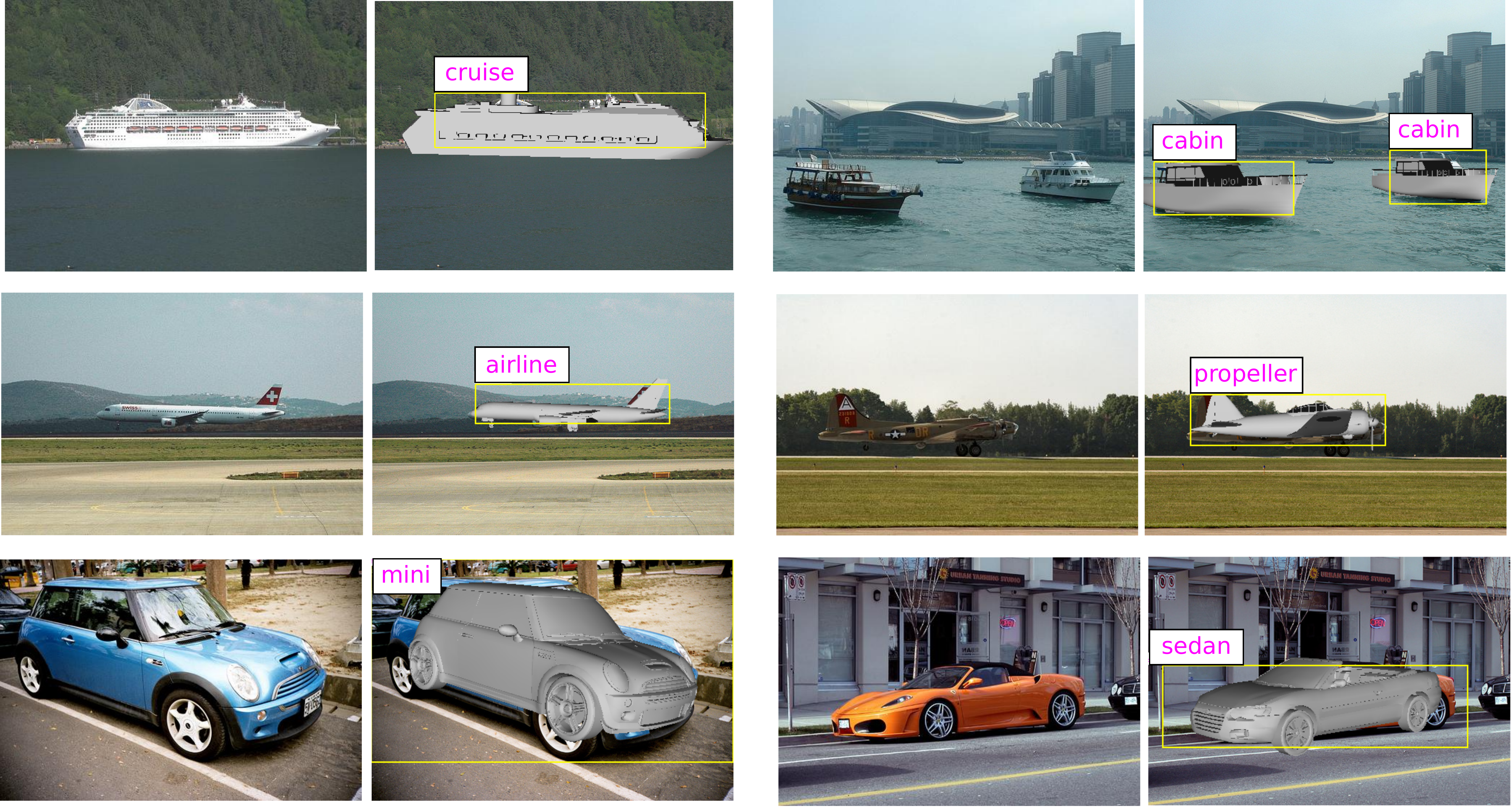}
  \caption{The result of object detection, 3D pose estimation, and (finer-)sub-category recognition. We show the projection of the 3D CAD model corresponding to the estimated finer-sub-categories according to the estimated continuous viewpoint. The magenta text is the estimated sub-category. Note that the 3D CAD model might not be the exact model for objects in PASCAL images.}  
  \label{fig:detres}
\end{figure*}

The result of our hierarchical model is shown in Table~\ref{tab:ours}. We consider three scenarios, a one-layer hierarchy, which is only the coarse viewpoint layer, a two-layer hierarchy, and a three-layer hierarchy, which is our full model. Unlike the DPM case, we typically do not observe a performance drop as we add more layers to the model. In some cases we see significant improvement. For instance, the result of sub-category recognition, and joint sub-category and viewpoint estimation improves by 3.1 and 1.6, respectively, for the 3-layer hierarchy compared to the 2-layer hierarchy. For detailed per-class results, refer to the supplementary material. 
\begin{figure*}
\centering
  \includegraphics[width=39pc]{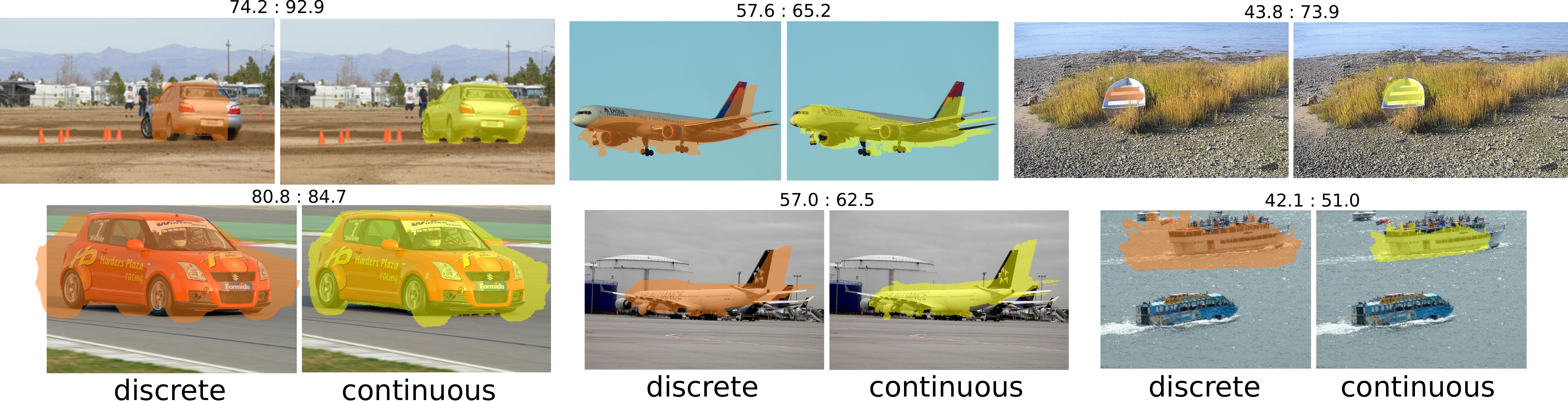}
  \caption{The left and the right image show the results of segmentation with the discrete and continuous versions of our model, respectively. The numbers on top are the corresponding intersection over union measures. Groundtruth segmentation mask is used to compute the overlap accuracy.\vspace{-0.5cm}}
  \label{fig:segres}
\end{figure*}

    \begin{table*}[!ht]
		    \vspace{-1.0cm}
        \begin{center}        
            \subtable
                {
                    {\scriptsize
                    \begin{tabular}{|l|c|c|}
                     \hline
		      \cellcolor{orr} CAD Alignment & 3-layer discrete & 3-layer continuous \\
		      \hline
		      aeroplane & 50.5 & \textbf{51.5} \\
		      \hline
		      boat & 35.7 & \textbf{40.3} \\
		      \hline
		      car & 60.4 & \textbf{64.4} \\		      		      
		      \hline                    
                    \end{tabular}\centering
                    }
                }
            \subtable
                {
                    {\scriptsize
                    \begin{tabular}{|l|c|c|}
                     \hline
		      \cellcolor{blue!45} 2D Segmentation & 3-layer discrete & 3-layer continuous \\
		      \hline
		      aeroplane & 36.5 & \textbf{37.4} \\
		      \hline
		      boat & 35.6 & \textbf{39.9} \\
		      \hline
		      car & 61.4 & \textbf{64.3} \\		      
		      \hline                
		      \end{tabular}\centering
		    }  
                }
            \caption{Segmentation results obtained by discrete and continuous versions of our model.}            
            \label{tab:segres}            
        \end{center}       
    \end{table*}

For the sake of comparison of viewpoint evaluations, we discretize the estimated continuous viewpoint into 8 azimuth angles. Note that the 1-layer hierarchy is already better than the current state-of-the-art (compare its results to DPM-VOC+VP \cite{pepik12a} in Table~\ref{tab:DPM}, which is the state-of-the-art in viewpoint estimation) partially because of the powerful CNN features. Therefore, providing improvement over the first layer is not an easy task. 
Also, note that the performance for `All' is quite low, which indicates the difficulty of modeling all tasks together. For instance, for \textit{cars}, in addition to object detection, we should correctly infer one of the 8 azimuth angles, one of the 8 sub-categories, and one of the $\sim8$ finer-sub-categories corresponding to the estimated sub-category. Figure~\ref{fig:detres} illustrates detection results for the 3-layer hierarchy. 

Note that more supervision should not necessarily result in better accuracy. The reason is that we consider more tasks (viewpoint, subcategory, etc.) to model as we increase supervision. As the number of tasks increases, the space of parameters becomes huge, and learning the optimal parameters becomes much harder than the case where we model only a single task. Mainly due to this issue, most works on joint object detection and 3D pose estimation (e.g., \cite{aubry14} or \cite{pepik12b}) are outperformed by DPM that uses less supervision for the single task of `bounding box detection'. Note however that DPM is not capable of 3D pose estimation.

In Table~\ref{tab:ours}, we also compare our hierarchical model to a flat model that uses the same set of features as those of the 3-layer hierarchy. The flat model is basically a linear classifier whose output labels are joint viewpoint and (finer-)sub-categories, and it is applied to the proposal regions. The confidence values we obtain by the flat model are different from those of the hierarchy, which results in large performance difference (the flat model is significantly lower). To compare viewpoint and subcategory estimation irrespective of the confidence, for the flat case, we consider the same confidence (energy) as that of the 3-layer hierarchy. As shown in the table, the 3-layer hierarchy provides significant improvement over the flat model. Even for the difficult `All' task we observe around 23\% improvement. Table~\ref{tab:ours} also includes the results for separate classifiers i.e., we have a classifier for viewpoint, a separate classifier for sub-category and another set of classifiers for finer-sub-categories (unlike the flat model that is a joint classifier).

We computed the RMSE for estimating azimuth, elevation and distance. The results are shown in Table~\ref{tab:contres}. Unfortunately, we cannot compare the results with other methods as other methods do not provide results for distance and elevation. We compare our method with \cite{pepik12a} for different discretizations of the azimuth in Table~\ref{tab:1624}. Note that our method is trained with 8 views. The confusion matrix for sub-category recognition for the \textit{car} category is shown in Figure~\ref{fig:carcm}. The confusion matrices for other categories can be found in the supplementary material. Note that the AVP measure favors dominant categories and we chose the parameters such that we maximize AVP. Hence, the confusion matrix is biased towards \textit{Sedan}, which is the dominant category. 

Note that DPM  \cite{felzenswalb10}, DPM-VOC-VP \cite{pepik12a}, or the flat model are classifiers for azimuth and it is impractical to incorporate other parameters of the continuous viewpoint into them since the output label space becomes huge. To show the advantage of our method that estimates continuous viewpoints over the discrete classifiers, we perform the following experiment. We project the CAD model corresponding to the estimated finer-sub-category according to the estimated continuous viewpoint and measure the intersection over union (IOU) of the projection mask with the groundtruth object mask. We consider two cases: 1) We use the projection of the groundtruth CAD given the groundtruth viewpoint as the groundtruth mask (referred to as `CAD Alignment' in Table~\ref{tab:segres}). 2) We use the groundtruth segmentation mask of \cite{berkeley11} for evaluation (referred to as `2D Segmentation'). Unlike case (1), this case considers occlusion by external objects as well. The result is shown in the right hand side of Table~\ref{tab:segres}. 
\begin{table}[t]
{\scriptsize
\begin{center}
 \begin{tabular}{|c|c|c|c|}
 \hline
 \cellcolor{green!45} RMSE & Azimuth (degree) & Elevation (degree) & Distance \\
 \hline
 Aeroplane & 73.15 & 19.21 & 8.19 \\
 \hline   
 Boat & 100.48 & 12.71 & 13.4 \\
 \hline   
 Car & 73.16 & 6.59 & 11.25\\
 \hline   
 \end{tabular}\centering
\end{center}
}
\caption{Continuous viewpoint estimation error.}
\label{tab:contres}
\end{table}

\begin{table}[t]
{\scriptsize
\begin{center}
 \begin{tabular}{|c|c|c|c|c|}
 \hline
 \cellcolor{red!45} AVP & 4 views & 8 views & 16 views & 24 views \\
 \hline
  \begin{tabular}[c]{@{}c@{}}3-layer hierarchy\\trained with 8 views\end{tabular} & \textbf{32.7} & \textbf{29.5} & 15.2 & 10.2 \\

 \hline   
 DPM-VOC+VP \cite{pepik12a} & 24.9 & 21.8 & \textbf{15.3} & \textbf{12.2} \\
 \hline   
 
 \end{tabular}\centering
\end{center}
}
\caption{Results for different discretization of azimuth.}
\label{tab:1624}
\end{table}

\begin{figure}[tp]
\centering
  \includegraphics[width=14pc]{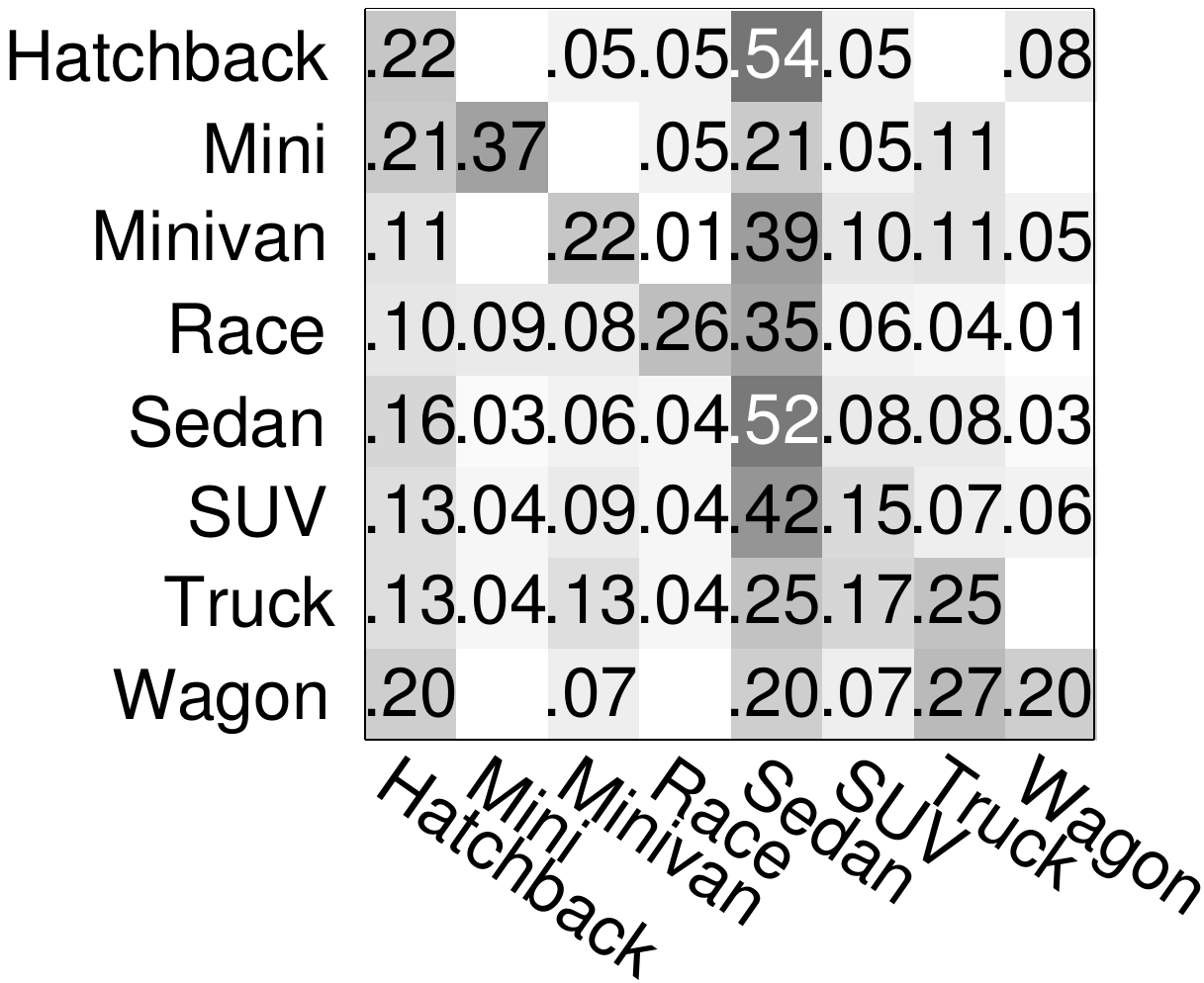}
  \caption{Confusion matrix for the sub-categories of the \textit{cars}.\vspace{-0.5cm}}  
  \label{fig:carcm}
\end{figure}


In both cases, using continuous viewpoint provides a significant improvement over the discrete case of our model (evaluated based on the standard PASCAL segmentation criteria), which means our continuous viewpoint provides better alignment with the objects. Note that for this evaluation we consider only the true positive bounding boxes. By `discrete version of our model', we mean the case that we ignore $\varphi_{cnt}$ in the model. For the discrete case, we assume the elevation is equal to the mean of the elevations in training data and the distance is equal to the distance of the sample with the highest weight (refer to the distance sampling procedure in Sec.~\ref{sec:pot}). Figure~\ref{fig:segres} shows some qualitative results.

\vspace{-0.3cm}

\section{Conclusion}
\vspace{-0.2cm}
We proposed a novel coarse-to-fine hierarchy as a unified framework for object detection, 3D pose estimation, and sub-category recognition. We showed that our hierarchical model is effective in modeling these tasks jointly. Additionally, we showed that continuous viewpoint estimation (which is not practical for discrete classifiers) provides better alignment with the groundtruth object and significantly improves segmentation accuracy. We  provided a new dataset that provides sub-category and finer-sub-category annotations for a subset of categories in PASCAL3D+ and used it to train and evaluate our model. 

\paragraph{Acknowledgments} We acknowledge the support of ONR grant N00014-13-1-0761 and NSF CAREER 1054127.

{\small
\bibliographystyle{ieee}
\bibliography{egbib}
}

\end{document}